\documentclass[10pt,twocolumn,letterpaper]{article}

\usepackage[pagenumbers]{iccv} 

\usepackage{adjustbox}

\definecolor{iccvblue}{rgb}{0.21,0.49,0.74}
\usepackage[pagebackref,breaklinks,colorlinks,allcolors=iccvblue]{hyperref}

\def\paperID{11696} 
\def\confName{ICCV}
\def\confYear{2025}

\title{Segment Any Mesh}

\author{George Tang\\
MIT\\
\and
William Zhao\\
MIT\\
\and
Logan Ford\\
Backflip AI\\
\and
David Benhaim\\
Backflip AI\\
\and 
Paul Zhang\\
MIT
}

\begin{document}
\twocolumn[{%
\maketitle
\begin{center}
    \captionsetup{type=figure}
    \includegraphics[width=\textwidth]{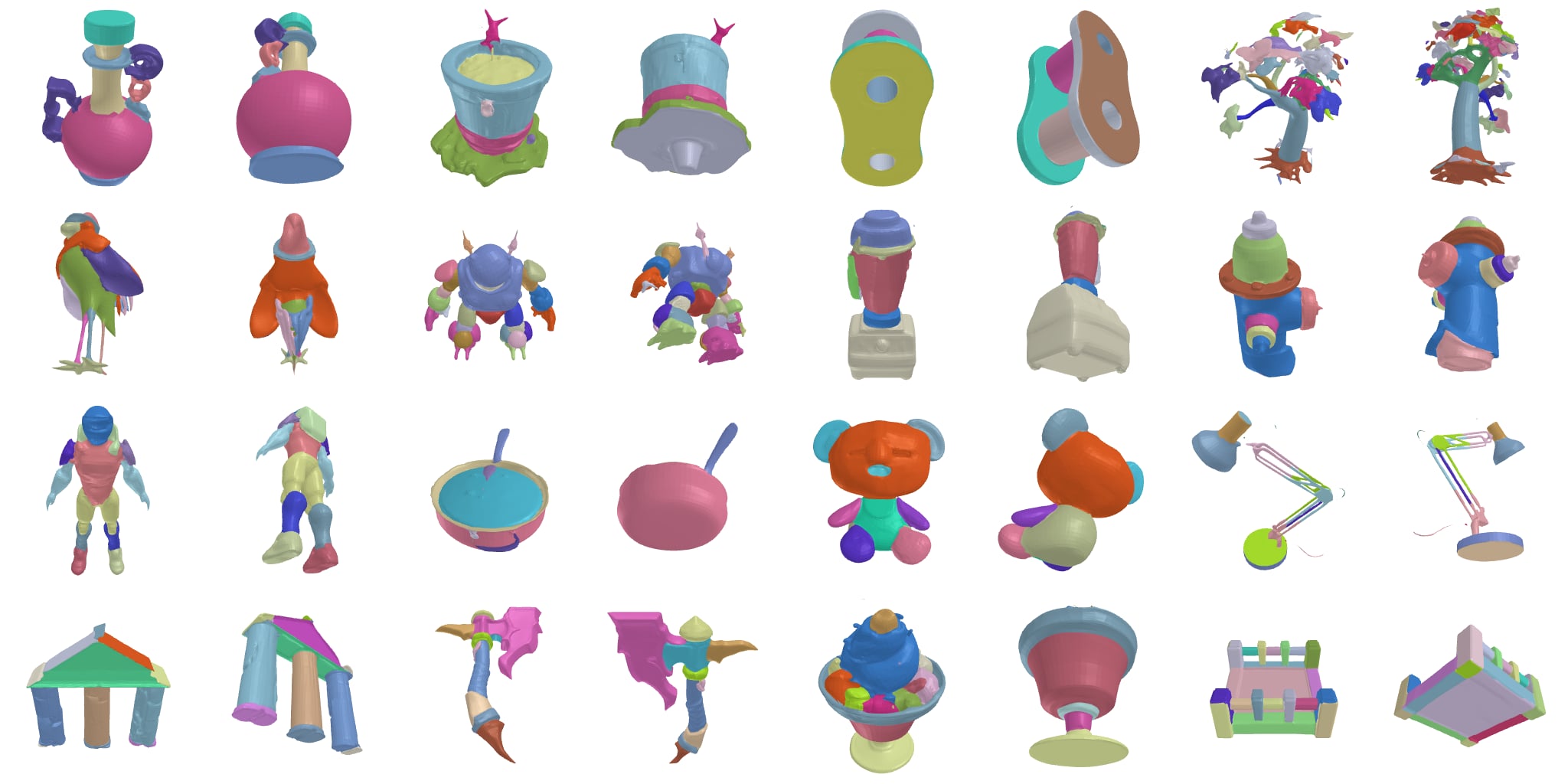}
    \captionof{figure}{Our method, Segment Any Mesh, lifts per view 2D segmentations into a mesh part segmentation. We show examples on our diverse object dataset curated from a 3D generative model. Segment Any Mesh generalizes well to novel shapes and object classes.}
    \label{fig:ours_backflip}
\end{center}
}]

\begin{abstract}
We propose Segment Any Mesh, a novel zero-shot mesh part segmentation method that overcomes the limitations of shape analysis-based, learning-based, and contemporary approaches. Our approach operates in two phases: multimodal rendering and 2D-to-3D lifting. In the first phase, multiview renders of the mesh are individually processed through Segment Anything to generate 2D masks. These masks are then lifted into a mesh part segmentation by associating masks that refer to the same mesh part across the multiview renders. We find that applying Segment Anything to multimodal feature renders of normals and shape diameter scalars achieves better results than using only untextured renders of meshes. By building our method on top of Segment Anything, we seamlessly inherit any future improvements made to 2D segmentation. We compare our method with a robust, well-evaluated shape analysis method, Shape Diameter Function, and show that our method is comparable to or exceeds its performance. Since current benchmarks contain limited object diversity, we also curate and release a dataset of generated meshes and use it to demonstrate our method's improved generalization over Shape Diameter Function via human evaluation.
\end{abstract}
\section{Introduction}
Mesh part segmentation has numerous applications ranging from texturing and quad-meshing in graphics to object understanding for robotics. Popular mesh part segmentation methods are either learning-based or employ more traditional shape analysis methods. For example, a robust shape analysis for mesh segmentation is Shape Diameter Function (ShapeDiam) \cite{Shapira2008ConsistentMP, Roy_2023}, which computes a scalar value per face that represents its local thickness. These values are clustered to get segmented regions.

However, learning-based methods are limited by the lack of diverse segmentation data \cite{abdelreheem2023satrzeroshotsemanticsegmentation, zhong2024meshsegmenterzeroshotmeshsemantic} while traditional methods do not work well beyond a limited number of mesh classes \cite{Chen:2009:ABF}. Recently, large 2D foundation models, such as Segment Anything (SAM) \cite{kirillov2023segment} and SAM2 \cite{ravi2024sam2segmentimages}, have achieved state-of-the-art results for image segmentation. This has revived interest in approaching 3D segmentation from aggregating masks produced using these 2D models on multiview renders of the mesh. Previous and contemporary 2D to 3D lifting methods for mesh segmentation, however, are restricted to semantic component segmentation as opposed to part segmentation since they require an input text description for the part to be segmented \cite{abdelreheem2023satrzeroshotsemanticsegmentation, zhong2024meshsegmenterzeroshotmeshsemantic}, making them unable to partition duplicate objects (e.g. arms, hands).

We propose Segment Any Mesh, a zero-shot approach for mesh part segmentation requiring only an input mesh. Our method consists of two phases, multimodal rendering and 2D to 3D lifting. During the first phase, renders from different angles \textit{individually} are fed into SAM to produce multiview masks\footnote{We utilize SAM2 for image segmentation, which produces better results than SAM. We do not employ video segmentation since the pose changes between views are larger than what SAM2 expects.}. In addition, we also render the IDs of the triangle faces of the mesh from each respective view. During the lifting phase, we use the multiview masks and face IDs to construct a match graph, which is used to associate 2D region labels with their corresponding 3D part. Specifically, we run community detection on this graph to obtain a rough mesh part segmentation, which we further postprocess to obtain the final segmentation.

\begin{figure*}[t]
    \centering
    \includegraphics[width=0.85\textwidth]{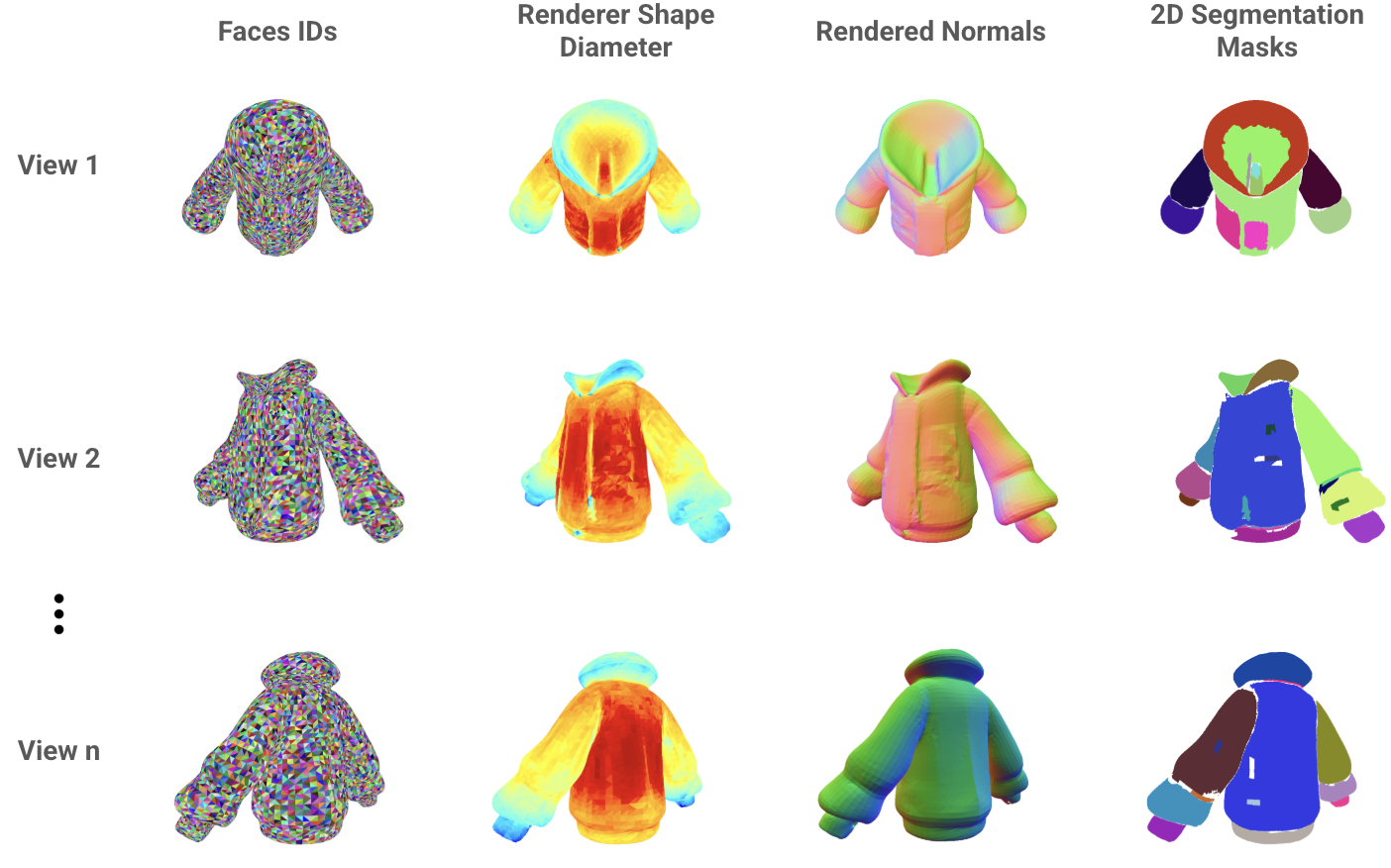}
    \caption{Our method applies Segment Anything to renderings of different modalities across views to get masks for each modality. These masks are then fused per view to get the per view 2D segmentation mask. FaceIDs, which are view consistent, are used to match segmentations across views.}
    \label{fig:method_meshes}
\end{figure*}

We operate in the untextured setting and find that untextured renderings do not result in SAM producing sufficiently detailed 2D masks that distinguish mesh parts. As an alternative, we find that SAM can operate on inputs from other modalities. Specifically, we show that by feeding both normal renders and ShapeDiam scalars renders of a geometry into SAM we can obtain detailed 2D masks with higher quality mesh part segmentations compared to relying on a single modality.

We benchmark our method by comparing it against ShapeDiam, which performs well on existing mesh part segmentation benchmarks. We show our method is comparable to or exceeds the performance of ShapeDiam on these benchmarks. Due to the limited object diversity of objects in these benchmarks, we curate and release a dataset of diverse object models created through a generative model. We show through human preference evaluation that the quality of segmentations produced by Segment Any Mesh greatly exceeds those produced by ShapeDiam.

To summarize, our contributions are
\begin{itemize}
    \item We propose Segment Any Mesh, a novel, zero-shot method that lifts masks produced by applying SAM on multiview renders into a 3D mesh part segmentation
    \item We show performance increases by working with masks fused from multimodal renders, specifically rendered surface normals and ShapeDiam scalars.
    \item We benchmark our method against the robust, well-evaluated Shape Diameter Function and demonstrate our method’s effectiveness on existing datasets. 
    \item Since these existing benchmarks do not exhibit much object diversity or complexity, we also curate a dataset from a custom 3D generative model and use it to demonstrate our method's generalization.
\end{itemize}
\section{Related Work}
\paragraph{Zero-Shot 2D Segmentation}
Advances in 2D image segmentation have been driven by the development of large-scale foundation models, which leverage extensive datasets and substantial compute \cite{oquab2024dinov2learningrobustvisual, liu2024groundingdinomarryingdino, cheng2022maskedattentionmasktransformeruniversal, zhang2023recognizeanythingstrongimage, liu2023visualinstructiontuning}. The ability to generalize across datasets without finetuning enables these models as powerful tools for application in downstream tasks. Among these models, Segment Anything Model (SAM) \cite{kirillov2023segment} and its successor SAM2 \cite{ravi2024sam2segmentimages} are currently state-of-the-art. SAM2 builds on SAM by improving the quality of segmentation and extends segmentation to dense video, though we do not leverage this capability in our method.

\paragraph{Zero-Shot 3D Segmentation}
In the realm of 3D segmentation, adjacent domains such as point cloud \cite{michele2023generativezeroshotlearningsemantic, chen2023zeroshotpointcloudsegmentation, liu2023partsliplowshotsegmentation3d}, and neural radiance fields (NeRFs) \cite{mildenhall2020nerfrepresentingscenesneural, zhi2021inplacescenelabellingunderstanding, tang2024efficient3dinstancemapping, cen2024segment3dradiancefields, kobayashi2022decomposingnerfeditingfeature}, have seen significant progress. However, mesh part segmentation is still at large. Learning-based methods for mesh segmentation \cite{Hanocka_2019, milano2020primaldualmeshconvolutionalneural, 10.1007/978-3-031-19818-2_31, smirnov2021hodgenetlearningspectralgeometry, koo2022partglotlearningshapesegmentation} are hindered by the limited availability of diverse segmentation datasets. Datasets like CoSeg \cite{10.1145/2366145.2366184} and Princeton Mesh Segmentation \cite{Chen:2009:ABF} contain only a few object classes, restricting the generalizability of models trained on them.

Shape analysis methods have long been used for 3D mesh segmentation. One robust method uses the Shape Diameter Function through the following steps: first, compute the ShapeDiam scalar for each face of the mesh (the local thickness of the shape); second, apply a one-dimensional Gaussian Mixture Model (GMM) to cluster these values into $k$ groups; third, perform an alpha expansion graph cut \cite{alphaexpansion} to segment the mesh based on these clusters; and finally, split disconnected regions with the same hierarchical label into distinct part labels. $k$ reflects the object part hierarchy rather than the number of object parts \cite{Shapira2008ConsistentMP, Roy_2023}. While effective in certain scenarios, these traditional methods are often limited in their ability to handle complex and diverse object classes, and they require careful parameter tuning to achieve optimal results.

Contemporary zero-shot 3D segmentation methods \cite{abdelreheem2023satrzeroshotsemanticsegmentation, abdelreheem2023zeroshot3dshapecorrespondence, zhong2024meshsegmenterzeroshotmeshsemantic, decatur20223dhighlighterlocalizingregions} are based on lifting the outputs of 2D foundation models and perform much better than previous multiview 3D segmentation \cite{su2015multiviewconvolutionalneuralnetworks, kalogerakis20173dshapesegmentationprojective} works due to a strong frontend 2D segmenter. However, they rely on input vocabulary and thus are limited to segmenting one part of an object at a time, effectively performing semantic rather than part segmentation. Furthermore, often parts of a mesh (e.g. CAD parts or components of architecture) do not have a suitable corresponding textual description. These approaches, while useful, do not fully address the need for automatic, comprehensive part segmentation in 3D meshes.
\section{Mesh Part Segmentation via 3D Lifting}
\begin{figure*}[t]
    \centering
    \includegraphics[width=\textwidth]{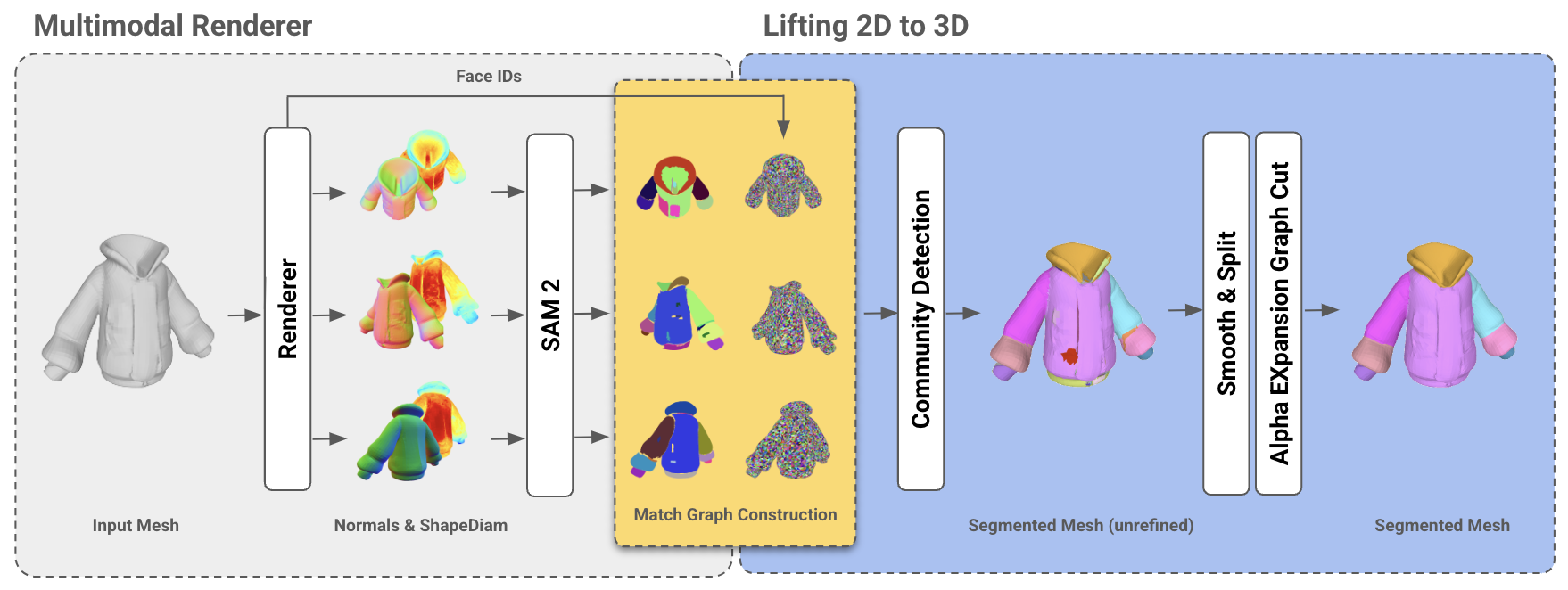}
    \caption{Overview of the Segment Any Mesh pipeline. In the Multimodal Renderer phase, surface normals and ShapeDiam scalars are rendered and fed into the SAM, and face IDs are also rendered. In the Lifting 2D to 3D phase, the match graph, which associates 2D labels that refer to the same mesh part, is constructed using the 2D masks and face IDs. Finally, community detection and postprocessing are applied to get the mesh part segmentation.}
    \label{fig:method}
\end{figure*}

\subsection{Multimodal Rendering}
In order to utilize SAM's powerful visual prior, we translate geometric information into visual information. Given a mesh $M$, we render $n$ views over varying modalities in a regular icosahedral layout around the given mesh. We then apply SAM to each render and, for each view, fuse masks from different modalities into a single 2D segmentation mask. 

We explore rendering three modalities and their combinations as inputs to SAM: untextured renderings, rendered surface normals, and rendered ShapeDiam scalars. We set SAM's prediction IOU threshold to 0.5 for all our experiments, lower than the default 0.8. We find that although this allows noisy masks to permeate each view, they are filtered out by the 2D to 3D lifting phase, while details are preserved. We also render face IDs, which are used for constructing the match graph in the next phase. Specifically, let $m_{i, \text{binary}}$ denote the binary masks accrued for view $i$, and the corresponding camera pose $p_i$. Let $\text{F}_j$ be the rendering function for modality $j$ for $J$ modalities, and $\odot$ be concatenation. Then $m_{i, \text{binary}}$ an be expressed as
\begin{align*}
     \text{SAM}(\text{F}_1(M, p_i)) \ \odot \ ... \ \odot \ \text{SAM}(\text{F}_J(M, p_i))
\end{align*}

Given more than one modality, we perform mask fusion. We sort all SAM binary masks from different modalities by area for each view. The masks are fused into a single instance segmentation mask, $m_{i,\text{instance}}$ by overlaying from the largest area binary masks on the bottom to smallest area binary masks on top (the mask area essentially defines a rasterization order for the binary masks). After mask overlaying, we remove islands and holes with an area less than $A_{\text{SAM}} = 1024$ pixels, which we use for all our experiments. Masks that fall in the background are removed using the face ID renders.

\subsection{Lifting Segment Anything}
We define our \emph{unweighted} match graph $G = (V, E)$, where each node corresponds to a 2D region label in a rendered view mask while an edge indicates if two nodes potentially refer to the same mesh part. For all pairs of 2D region labels $(r_1, r_2) \in E$ across all $m_{i,\text{instance}}$, we calculate the overlap ratio of their 2D masks, $R$, to determine whether an edge should be added between their respective nodes. Specifically, let $\text{OF}(r_1)$ be the number of faces the projections of $r_1$ and $r_2$ onto the mesh share and $\text{F}(r)$ be the number of faces $r$’s projection occupies. If
\begin{align*}
\min\left(\frac{\text{OF}(r_1, r_2)}{\text{F}(r_1)}, \frac{\text{OF}(r_1, r_2)}{\text{F}(r_2)} \right) > \tau_R
\end{align*}

and the condition $\text{OF}(r_1, r_2) > \tau_C$, where $\tau_C$ is an overlap threshold to filter noise, we connect nodes $r_1$ and $r_2$. We set $\tau_C = 32$ in all our experiments. We observe that the optimal $\tau_R$ for different meshes can vary but a somewhat decent threshold can be approximated dynamically as follows: we discretize the overlap ratio $R \in [0, 1]$ over all region pairs into a histogram $H$ with resolution $r_H = 100$ bins. We set $\tau_R$ to the bin where the prefix sum of the number of edge candidates is at least fraction $p_{\tau_R}$ of the total number of edge candidates. We vary $p_{\tau_R}$ in our experiments depending on the dataset (see Section Implementation).
\begin{align*}
    \tau_R &= \min \left( b : \sum_i^b H(i) > p_{\tau_R}N_{\text{pairs}}\right)
\end{align*}

We apply Leiden community detection on the match graph using \texttt{igraph}'s implementation with resolution parameter 0. Afterwards, we filter out communities with size $\tau_{CD} = 1$, to get node communities. The resulting nodes’ labels in each respective community refer to the same mesh part segmentation label. We then project the mesh part segmentation labels onto the faces, with the canonical part segmentation label per face set as the most referred label. Ties are broken arbitrarily, though we observe they are negligible in quantity.

\subsection{Mesh Segmentation Refinement}
We first remove holes (unlabeled space) with an area less than $A_{\text{mesh}} = p_{A_{\text{mesh}}} \cdot N_{\text{faces}}$, where the face count fraction $p_{A_{\text{mesh}}}$ is 0.025 in all experiments. We then handle islands by expanding the frontier by one face each iteration for $I_{\text{smooth}}$ iterations. We set $I_{\text{smooth}}$ to a large number (64 in all experiments) so that the remaining mesh has no holes. We then split disconnected regions and, as in ShapeDiam follow up by applying an alpha expansion graph cut \cite{alphaexpansion} for 1 iteration\footnote{For all our experiments, we use a smoothing value of $\alpha=4$ as done in \cite{Shapira2008ConsistentMP}}, which we use in all our experiments. We weight the alpha expansion graph cut cost term by a factor of $\lambda$, which we vary in our experiments depending on the dataset (see section Implementation).

\section{Experiments}
\begin{figure*}[!htp]
    \centering
    \includegraphics[width=0.95\textwidth]{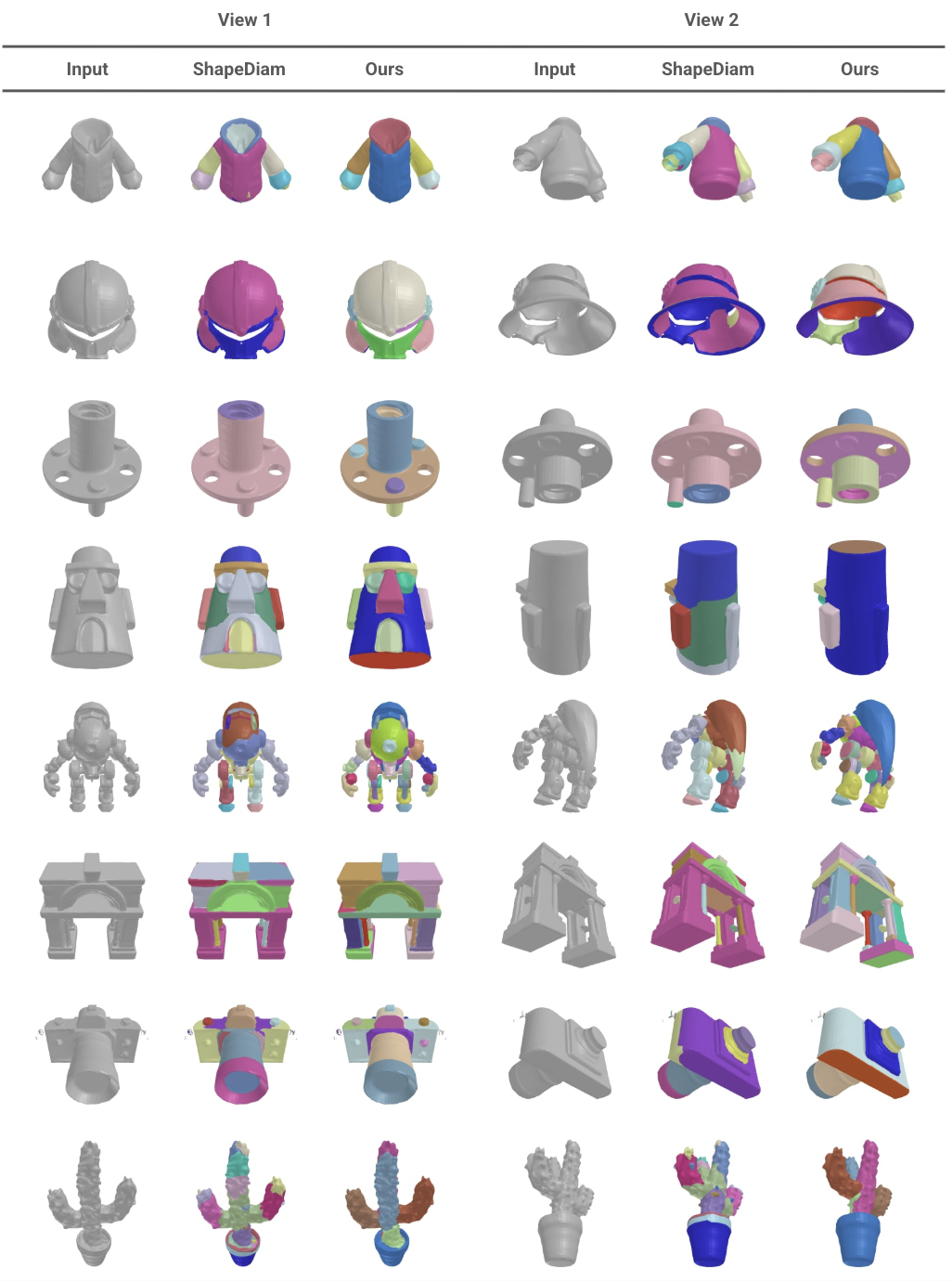}
    \caption{We show comparisons between Segment Any Mesh and ShapeDiam for meshes from our curated dataset.}
    \label{fig:ours_backflip_vs_shapediam}
\end{figure*}

\paragraph{Datasets}
We first conduct a human evaluation study for Segment Any Mesh vs ShapeDiam as well as ablations on the modalities used. We curate a diverse set of meshes from a custom 3D generative model consisting of 75 diverse watertight meshes— daily objects, architecture, and artistic objects. We release the dataset to the public domain.

For the human evaluation protocol, we concatenate rendered videos of the segmented meshes color mapped with the same set of visually distinct colors, from largest region area to smallest region area, and ask the human evaluator to rank the videos. The order of videos in the concatenation was randomized across meshes. In addition, the ablations of the modalities fed to SAM included untextured renderings, rendered surface normals, rendered ShapeDiam scalars, and combining rendered surface normals and rendered ShapeDiam scalars. We do not benchmark on all combinatorially possible cases since 1) we do not notice much improvement using untextured renderings as a modality compared to rendered surface normals and ShapeDiam scalars, and 2) the cognitive load of ranking more than 4 concatenated videos will lead to inaccurate evaluation.

We separately compare our method against ShapeDiam on traditional mesh segmentation benchmark datasets, in which ShapeDiam achieves decent results and represents the upper ceiling of previous nonlearning approaches. We benchmark on the CoSeg dataset and the Princeton Mesh Segmentation Dataset against human-annotated ground truth. Unlike our curated dataset, the CoSeg dataset is composed of 8 classes of objects while the Princeton Mesh Segmentation has 19, and they are more useful for measuring segmentation consistency for meshes belonging same class as opposed to generalization.

\paragraph{Metrics}
For the Human Evaluation Study, each mesh had $n = 5$ trials, and we computed the mean/std rank (starting at 1) across all trials and meshes for each method. 

For CoSeg and Princeton Mesh Segmentation, we use the 7 metrics introduced in Princeton Mesh Segmentation. We briefly introduce them below.

\begin{enumerate}
    \item \emph{Cut Discrepancy} ($\downarrow$) sums the distances from points along the cuts in the generated segmentation to the nearest cuts in the ground truth segmentation, and vice versa. This provides an intuitive measure of the separation between the cuts.
    \item \emph{Hamming Distance} ($\downarrow$) measures the overall region-based difference between two segmentations.
    \item \emph{Hamming Distance (Rf)} ($\downarrow$) assuming second segmentation is ground truth, measures false positive labels
    \item \emph{Hamming Distance (Rm)} ($\downarrow$) assuming second segmentation is ground truth, measures true negative labels
    \item \emph{Rand Index} measures the likelihood that a pair of faces are either grouped within the same segment or separated into different segments across two segmentations.
    \item \emph{Local Consistency Error} ($\downarrow$) accounts for nested, hierarchical similarities in segmentations
    \item \emph{Global Consistency Error} ($\downarrow$) accounts for nested, hierarchical similarities in segmentations, forcing all local refinements to be in the same direction
\end{enumerate}

For exact formulations, please reference \cite{Chen:2009:ABF}.

\begin{figure*}[!t]
    \centering
    \includegraphics[width=0.85\textwidth]{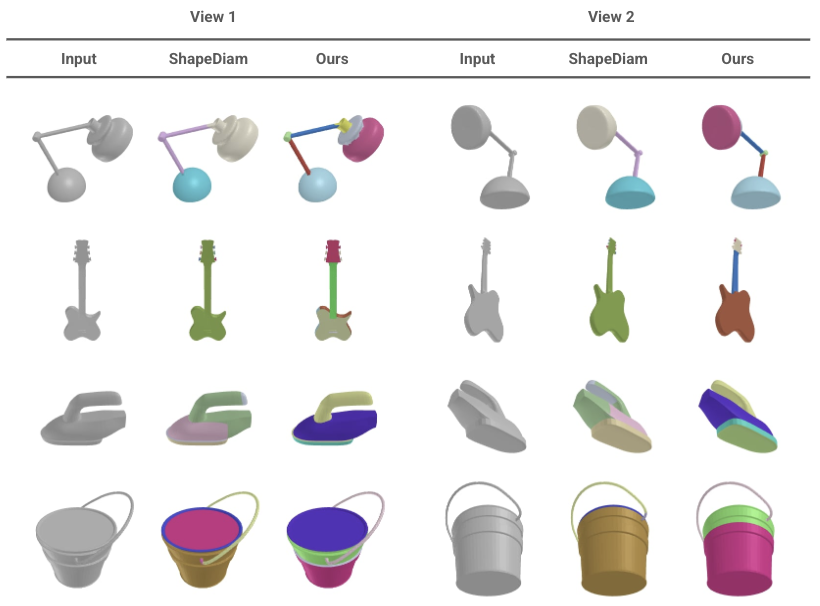}
    \caption{Qualitative examples of Segment Any Mesh vs ShapeDiam outputs on the CoSeg dataset. We provide additional comparisons with ShapeDiam in the supplemental.}
    \label{fig:ours_coseg}
\end{figure*}

\paragraph{Implementation}
We run all experiments on a single Nvidia A10 GPU. For our human evaluation study, we recruited a community of $m=17$ participants ($n=5$ trials per mesh). The workforce was asked to rank the segmentations from best to worst. We show the interface and evaluation criteria in the supplementary material. For Princeton Mesh Segmentation Evaluation, we use the provided segmented meshes for ShapeDiam while we implement our own ShapeDiam for use on the other two datasets.

We now discuss how we choose our parameters. Larger $\lambda$ means more smoothing and fewer components, which is desirable for coarse segmentations. Similarly, lower $\tau_R$ correlates with fewer edges being added to the match graph and fewer components. For the human evaluation study and Princeton Mesh Segmentation Benchmark, we set Segment Any Mesh's $\lambda = 6$ and $\tau_R = 0.125$ for human evaluation and $\tau_r = 0.05$ for CoSeg. For ShapeDiam, we observe the best performance with $k=5$ GMM clusters and $\lambda = 15$ for human evaluation and Princeton Mesh Segmentation Benchmark, while for CoSeg, $k=3$ (see CoSeg dataset visualizations on its webpage) and $\lambda = 15$.

\begin{figure}[h]
    \centering
    \includegraphics[width=0.45\textwidth]{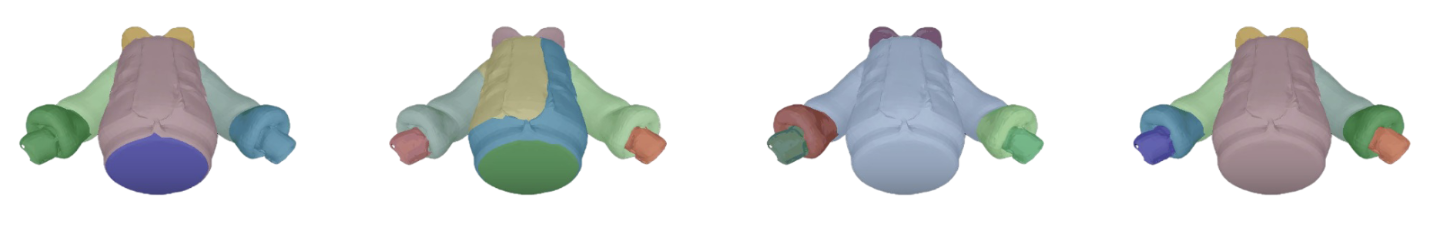}
    \caption{Comparison of mesh part segmentations produced by varying input modalities to Segment Anything. From left to right, rendered surface normals only, rendered ShapeDiam scalars only, untextured rendering, and rendered surface normals combined with rendered ShapeDiam scalars (best).}
    \label{fig:method_ablation.png}
\end{figure}

\paragraph{Human Evaluation Study}
Table \ref{tab:ours_backflip_vs_shapediam} shows the mean and standard deviation of human rankings for Segment Any Mesh vs ShapeDiam. We can see that Segment Any Mesh consistently ranks above ShapeDiam. Qualitative results for Segment Any Mesh on the generated dataset are shown in Figure \ref{fig:ours_backflip} while comparisons between Segment Any Mesh and ShapeDiam are shown in Figure \ref{fig:ours_backflip_vs_shapediam}. Table \ref{tab:method_ablation} shows the mean and standard deviation for rankings of ablations on input modalities to SAM, we see that combining surface normal and ShapeDiam scalar value renderings are evaluated as the best. Figure \ref{fig:method_ablation.png} shows a comparison between the mesh part segmentations produced by using different input modalities.

\begin{table}[!t]
    \centering
    \adjustbox{max width=\linewidth}{
    \begin{tabular}{lrr}
    \toprule
        Model & Rank Mean ($\downarrow$) & Rank Std ($\downarrow$) \\
    \midrule
        ShapeDiam & 1.82 & 0.27 \\
        Ours & \textbf{1.18} & 0.27 \\
    \bottomrule
    \end{tabular}
    } 
    \caption{Human evaluation rankings for Segment Any Mesh vs ShapeDiam.}
    \label{tab:ours_backflip_vs_shapediam}
\end{table}

\begin{table}[!t]
    \centering
    \adjustbox{max width=\linewidth}{
    \begin{tabular}{lrr}
    \toprule
        Rendering Modality & Rank Mean ($\downarrow$) & Rank Std ($\downarrow$) \\
    \midrule
        Ours (only norm) & 2.57 & 0.58 \\ 
        Ours (only shape diameter function) & 2.52 & 0.75 \\ 
        Ours (untextured) & 2.59 & 0.70 \\
        Ours & \textbf{2.33} & 0.59 \\
    \bottomrule
    \end{tabular}
    } 
    \caption{Human evaluation rankings for different modality combinations used as input to Segment Anything. We only perform 4 ablations as any number beyond that results in a ranking task that is too difficult.}
    \label{tab:method_ablation}
\end{table}

\begin{figure*}[!t]
    \centering
    \includegraphics[width=0.85\textwidth]{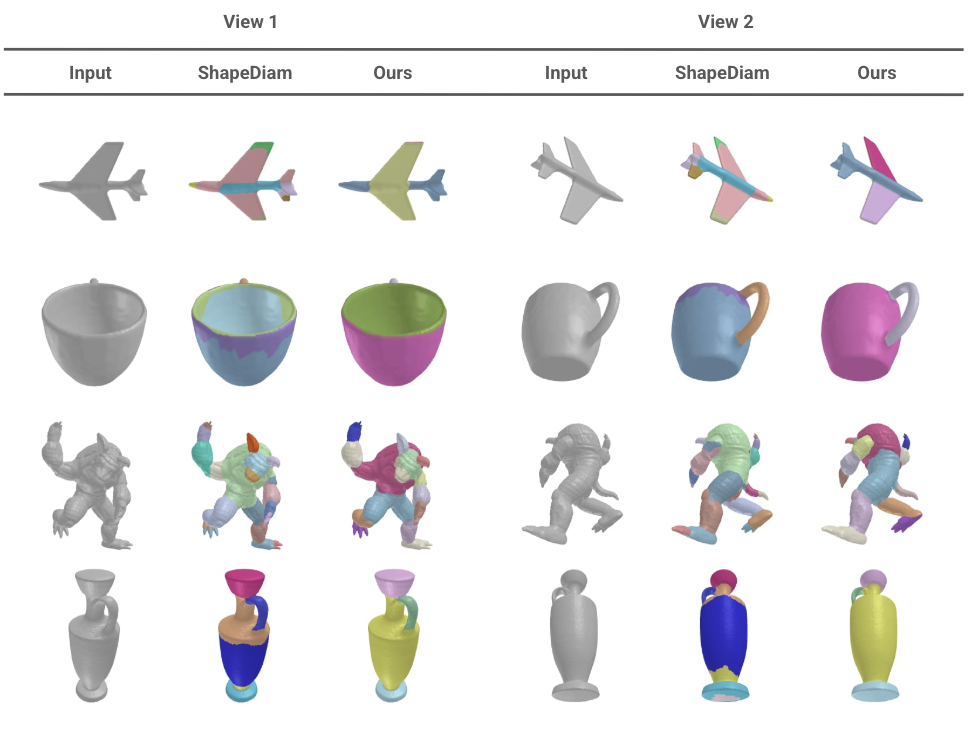}
    \caption{Qualitative examples of Segment Any Mesh vs ShapeDiam outputs on the Princeton Mesh Segmentation Benchmark dataset. We provide additional comparisons with ShapeDiam in the supplemental.}
    \label{fig:ours_princeton}
\end{figure*}

\paragraph{Traditional Benchmarks}
We now show Segment Any Mesh performs as well as ShapeDiam on traditional benchmarks, which ShapeDiam is overfitted to. Note that metrics should only be taken as an estimate of segmentation quality since mesh part segmentation can be quite subjective. Table \ref{tab:ours_coseg} shows we match the performance of ShapeDiam, and Figure \ref{fig:ours_coseg} shows example outputs of our method on the CoSeg dataset. Table \ref{tab:ours_princeton} shows we also exceed the performance of ShapeDiam, and Figure \ref{fig:ours_princeton} shows example outputs of our method on the Princeton Mesh Segmentation benchmark.

\begin{table}[h]
    \centering
    \adjustbox{max width=\linewidth}{
    \begin{tabular}{lrr}
    \toprule
        Metric & ShapeDiam & Ours \\
    \midrule
        Cut Discrepancy & 0.39 & \textbf{0.37} \\ 
        Hamming Distance & \textbf{0.16} & 0.18 \\
        Hamming Distance (Rf) & 0.10 & \textbf{0.08} \\ 
        Hamming Distance (Rm) & \textbf{0.23} & 0.28 \\ 
        Rand Index & \textbf{0.21} & 0.22 \\ 
        Local Consistency Error & 0.06 & \textbf{0.05} \\ 
        Global Consistency Error & 0.09 & \textbf{0.08} \\
    \bottomrule
    \end{tabular}
    } 
    \caption{Quantitative results for ShapeDiam vs Segment Any Mesh on the CoSeg dataset for automatic determination of the number of segmentation regions.}
    \label{tab:ours_coseg}
\end{table}

\begin{table}[h]
    \centering
    \adjustbox{max width=\linewidth}{
    \begin{tabular}{lrr}
    \toprule
        Metric & ShapeDiam & Ours \\
    \midrule
        Cut Discrepancy & 0.34 & \textbf{0.31} \\ 
        Hamming Distance & 0.21 & \textbf{0.17} \\ 
        Hamming Distance (Rf) & 0.17 & 0.17 \\ 
        Hamming Distance (Rm) & 0.26 & \textbf{0.17} \\ 
        Rand Index & 0.22 & \textbf{0.21} \\ 
        Local Consistency Error & 0.10 & \textbf{0.07} \\ 
        Global Consistency Error & 0.16 & \textbf{0.12} \\
    \bottomrule
    \end{tabular}
    } 
    \caption{Quantitative results for Princeton Mesh Segmentation Benchmark for ShapeDiam vs Segment Any Mesh for automatic determination of the number of segmentation regions.}
    \label{tab:ours_princeton}
\end{table}

Additionally, the Princeton Mesh Segmentation Benchmark provides mesh segmentations for ShapeDiam that use the mode number of human segmentation regions as the target number of GMM clusters, $k$. We benchmark our method against these mesh segmentations to reflect the scenario when the target number of segmentation regions is roughly known. Specifically, we search $\lambda$—the weight of the cost for alpha graph expansion. We select the smallest $\lambda$ that results in the number of segments produced by our method within a predefined error margin of the mode value. We set $\tau_R = 0.35$ while $\lambda$ is variable in the range $[1, 15]$. Table \ref{tab:ours_princeton_dynamic} shows our method matches the performance of ShapeDiam.

\begin{table}[h]
    \centering
    \adjustbox{max width=\linewidth}{
    \begin{tabular}{lrr}
    \toprule
        Metric & ShapeDiam & Ours \\
    \midrule
        Cut Discrepancy & \textbf{0.28} & 0.30 \\ 
        Hamming Distance & \textbf{0.16} & 0.17 \\ 
        Hamming Distance (Rf) & 0.14 & 0.14 \\ 
        Hamming Distance (Rm) & \textbf{0.18} & 0.20 \\ 
        Rand Index & \textbf{0.17} & 0.18 \\ 
        Local Consistency Error & 0.08 & 0.08 \\ 
        Global Consistency Error & 0.13 & 0.13 \\
    \bottomrule
    \end{tabular}
    } 
    \caption{Quantitative results for Princeton Mesh Segmentation Benchmark for ShapeDiam vs Segment Any Mesh given a reference number of segmentation regions determined by labelers.}
    \label{tab:ours_princeton_dynamic}
\end{table}
\section{Conclusion}
Segment Any Mesh performs well on existing benchmarks and generalizes for diverse classes of meshes. This is due to our method being based on lifting rather than learning from limited diversity segmentation data or local shape descriptors. Lifting tackles the problem from a 2D view perspective, aligning with humans when analyzing object parts and affordances. 

In the future, an interface can be developed refining the segmentations, which can further improve quality for targeted tasks such as object understanding, texturing, and retopology. We can also leverage this human-in-the-loop process to curate and distill a labeled segmentation dataset into a MeshCNN model \cite{Hanocka_2019}.
{
    \small
    \bibliographystyle{ieeenat_fullname}
    \bibliography{main}
}

\clearpage

\pagenumbering{arabic}
\renewcommand*{\thepage}{A\arabic{page}}

\setcounter{figure}{0}
\renewcommand\thefigure{S\arabic{figure}}   

\setcounter{table}{0}
\renewcommand\thetable{S\arabic{table}}   

\onecolumn
\begin{center}
    {\Large \textbf{Supplementary Material}}\\[10mm]
\end{center}

\appendix

\section{Additional Details for Human Evaluation Study}
We use AWS Sagemaker as an evaluation interface. For the Segment Any Mesh versus shape diameter function comparison, we render a video of the meshes side by side, randomized, and ask the participants to rank which segmentations they prefer in order of (a), (b), (c), (d) where the letter corresponds to the respective segmentation index in the video. Below are the interfaces participants interacted with

\begin{figure}[htp!]
    \centering
    \includegraphics[width=0.95\textwidth]{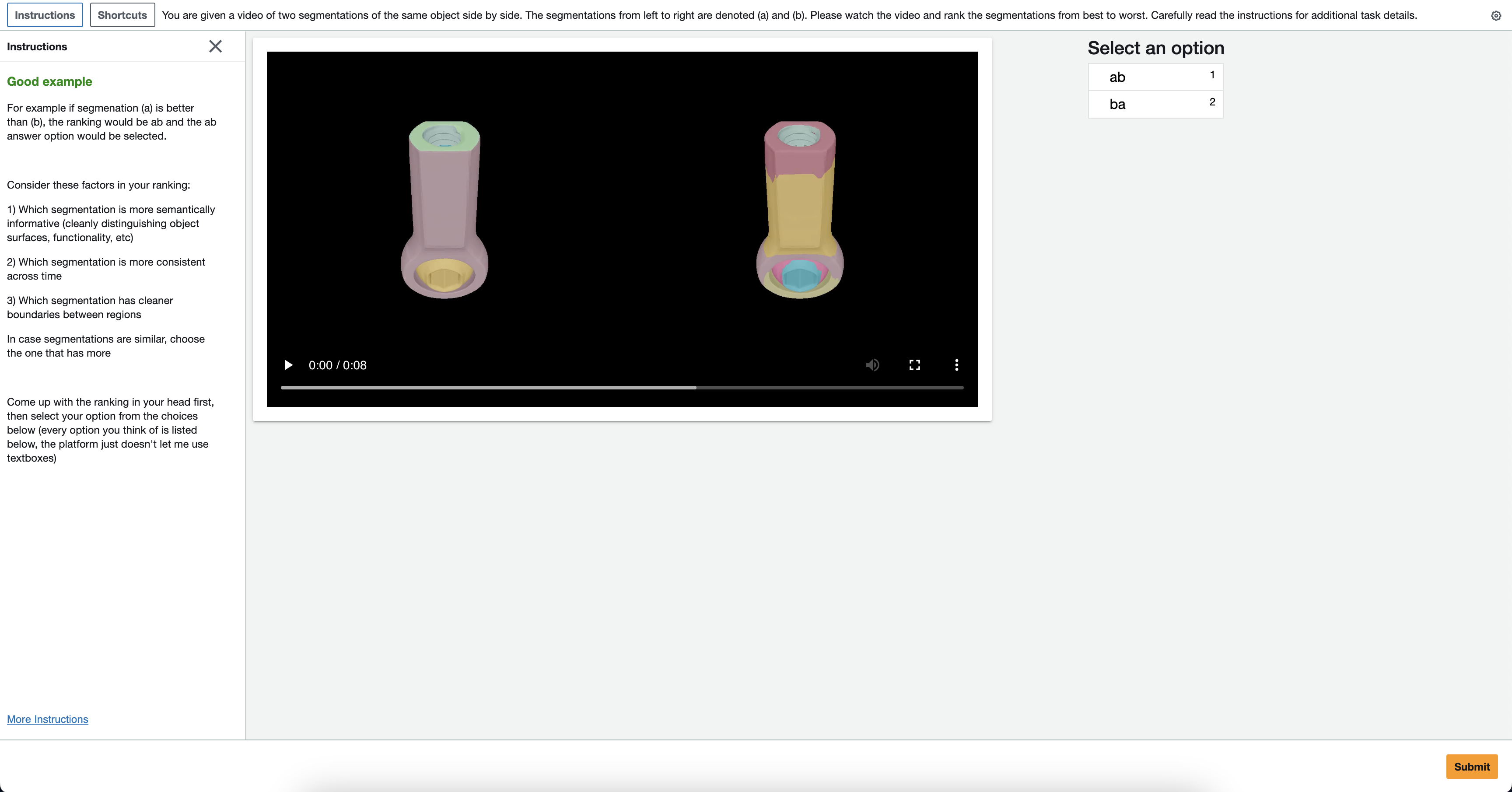}
    \caption{Interace for Segment Any Mesh vs Shape Diameter Function experiment.}
    \label{fig:ours_vs_shapediam_interface}
\end{figure}

\begin{figure}[htp!]
    \centering
    \includegraphics[width=0.95\textwidth]{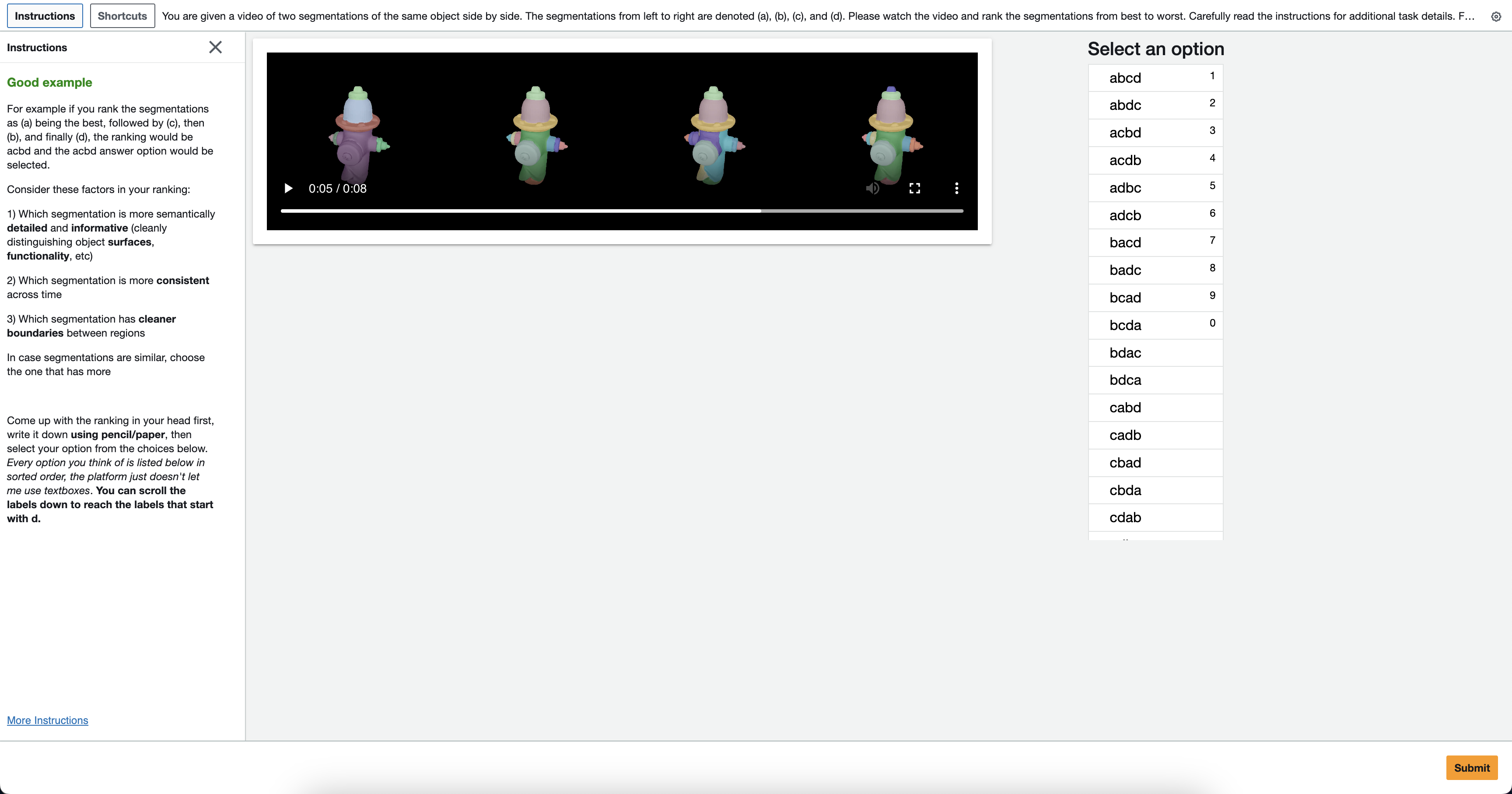}
    \caption{Interface for Segment Any Mesh vs Shape Diameter Function experiment. AWS Sagemaker does not support text input, so we resort to 24 labels corresponding to the ranking permutations.}
    \label{fig:ablation_modalities_interface}
\end{figure}

\newpage
We also include examples of frames from videos shown to the participants
\begin{figure}[htp!]
    \centering
    \includegraphics[width=0.75\textwidth]{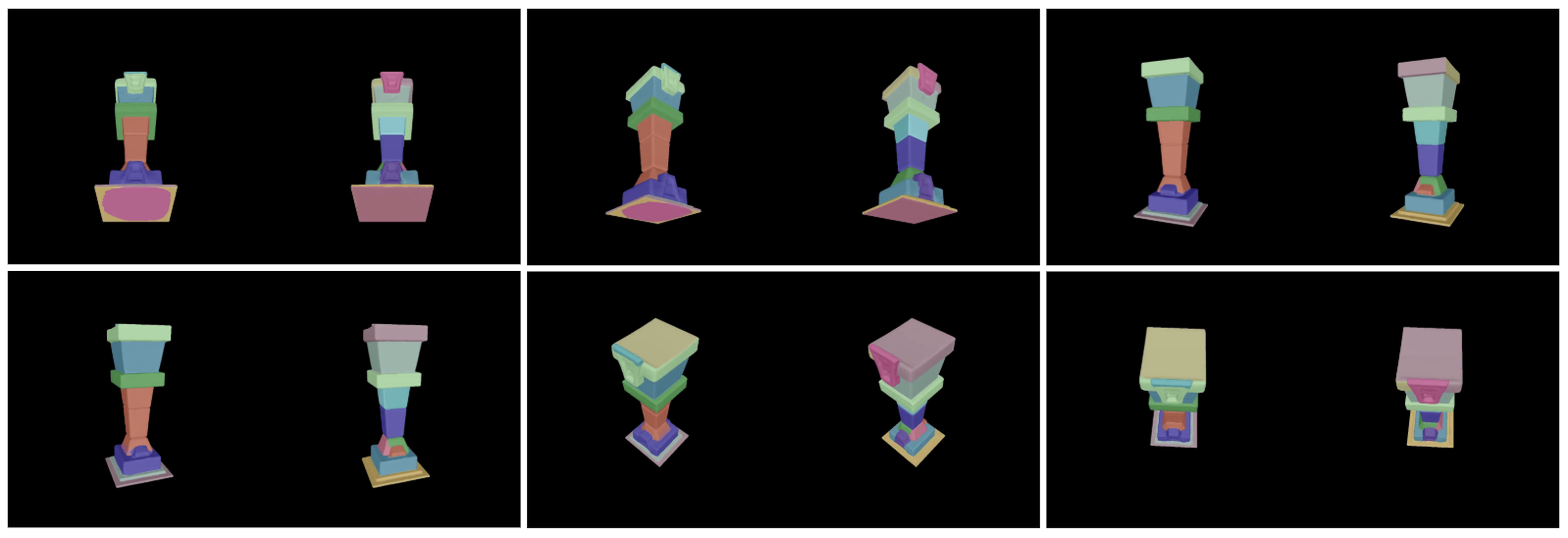}
    \caption{Frames from a video shown to participants for Segment Any Mesh vs Shape Diameter Function preference experiment.}
    \label{fig:ours_vs_shapediam}
\end{figure}

\begin{figure}[htp!]
    \centering
    \includegraphics[width=0.45\textwidth]{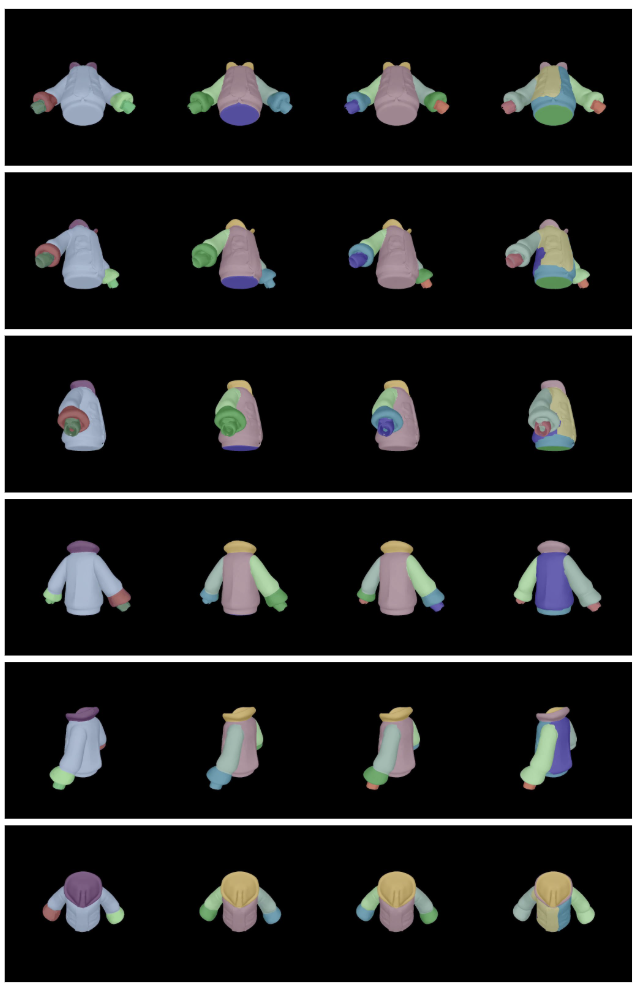}
    \caption{Frames from a video shown to participants for SAM input modality ablation experiment.}
    \label{fig:ablation_modalities}
\end{figure}

\section{Additional Visualizations}
We also provide Segment Any Mesh vs Shape Diameter Function comparisons for CoSeg and Princeton Mesh Segmentation Benchmark.

\newpage
\begin{figure}[htp!]
    \centering
    \includegraphics[width=0.95\textwidth]{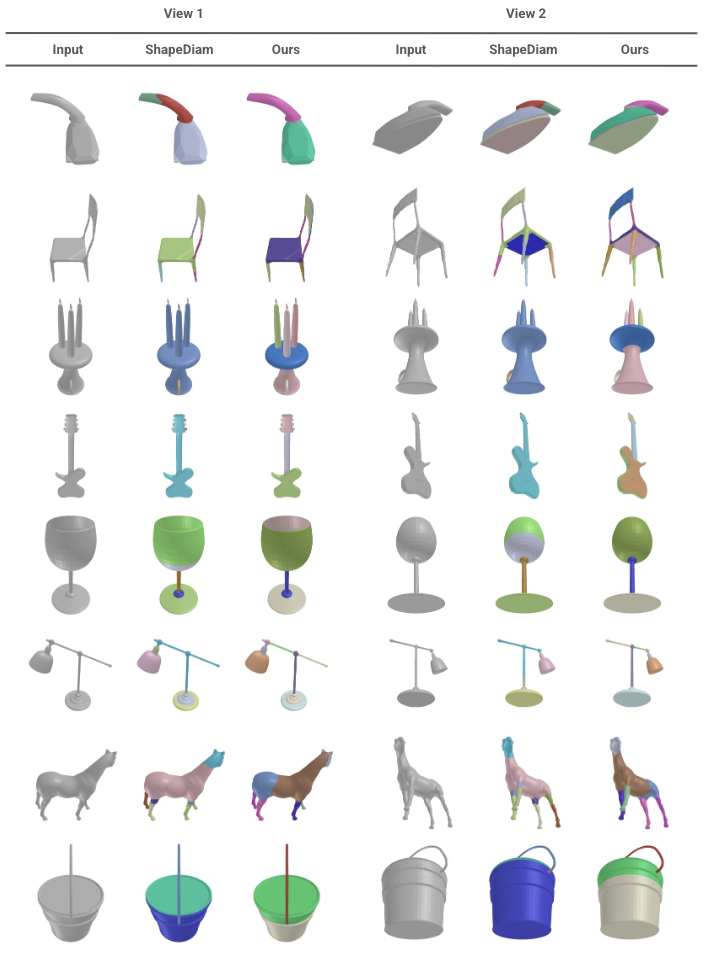}
    \caption{Segment Any Mesh vs Shape Diameter Function on the CoSeg dataset.}
    \label{fig:ours_vs_shapediam_coseg}
\end{figure}

\begin{figure}[htp!]
    \centering
    \includegraphics[width=0.95\textwidth]{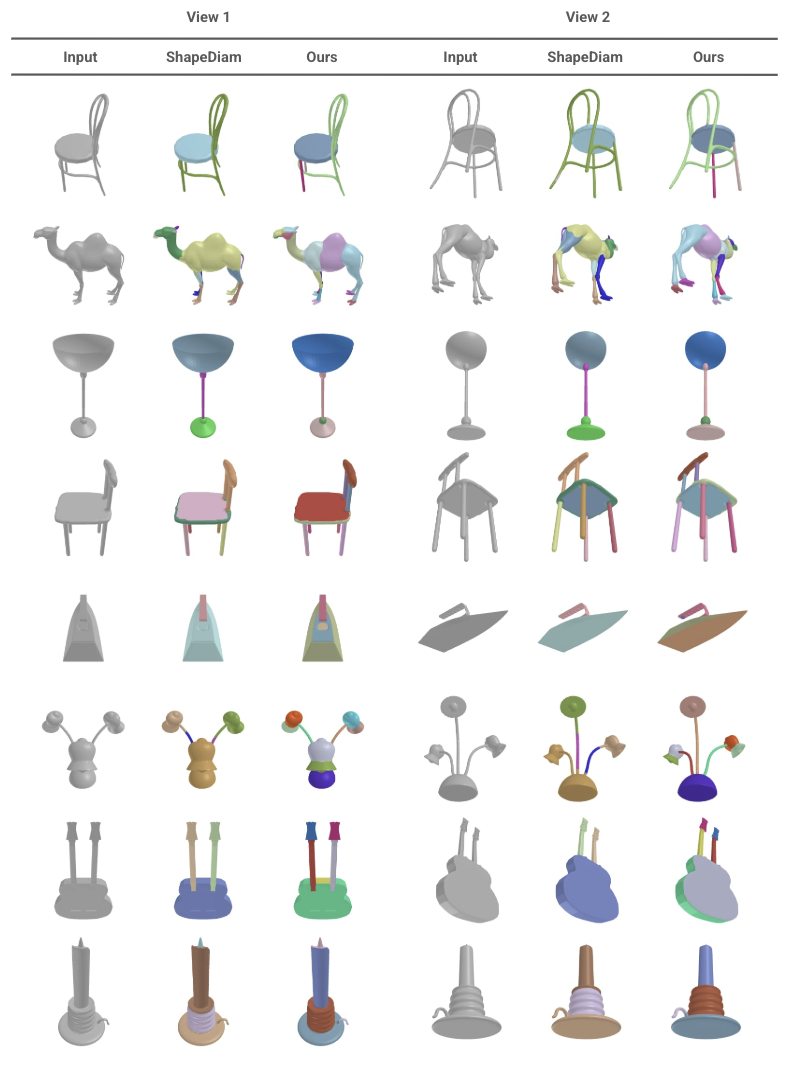}
    \caption{Additional visualizations for Segment Any Mesh vs Shape Diameter Function on the CoSeg dataset.}
    \label{fig:ours_vs_shapediam_coseg2}
\end{figure}

\begin{figure}[htp!]
    \centering
    \includegraphics[width=0.95\textwidth]{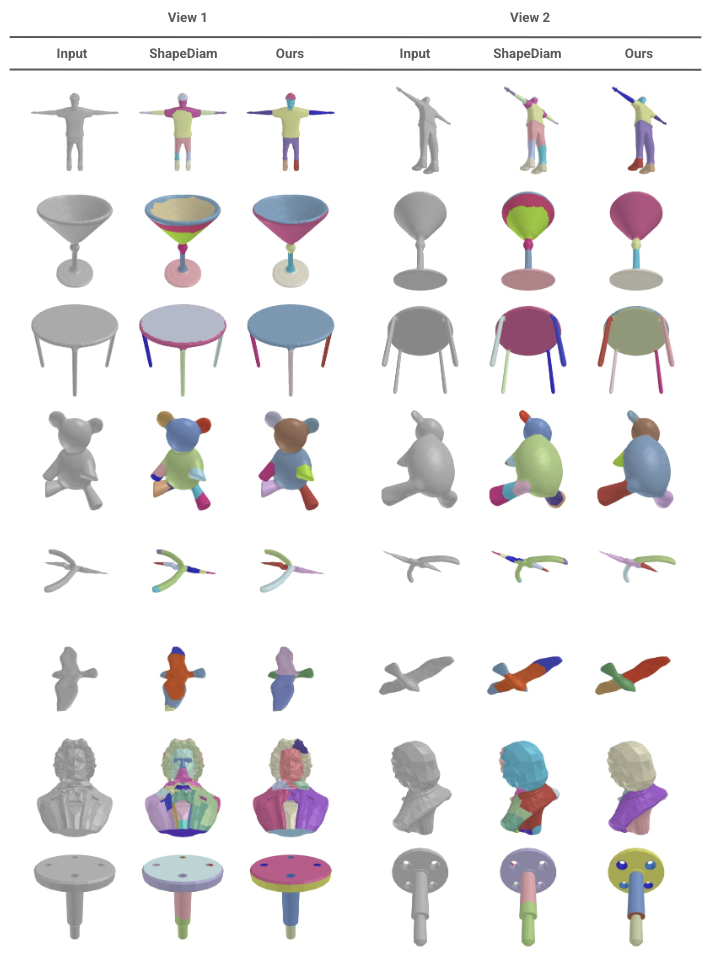}
    \caption{Segment Any Mesh vs Shape Diameter Function on the Princeton Mesh Segmentation Benchmark for automatic determination of the number of segmentation regions i.e. not given the reference number of segmentation regions $k$.}
    \label{fig:ours_vs_shapediam_princeton}
\end{figure}

\begin{figure}[htp!]
    \centering
    \includegraphics[width=0.95\textwidth]{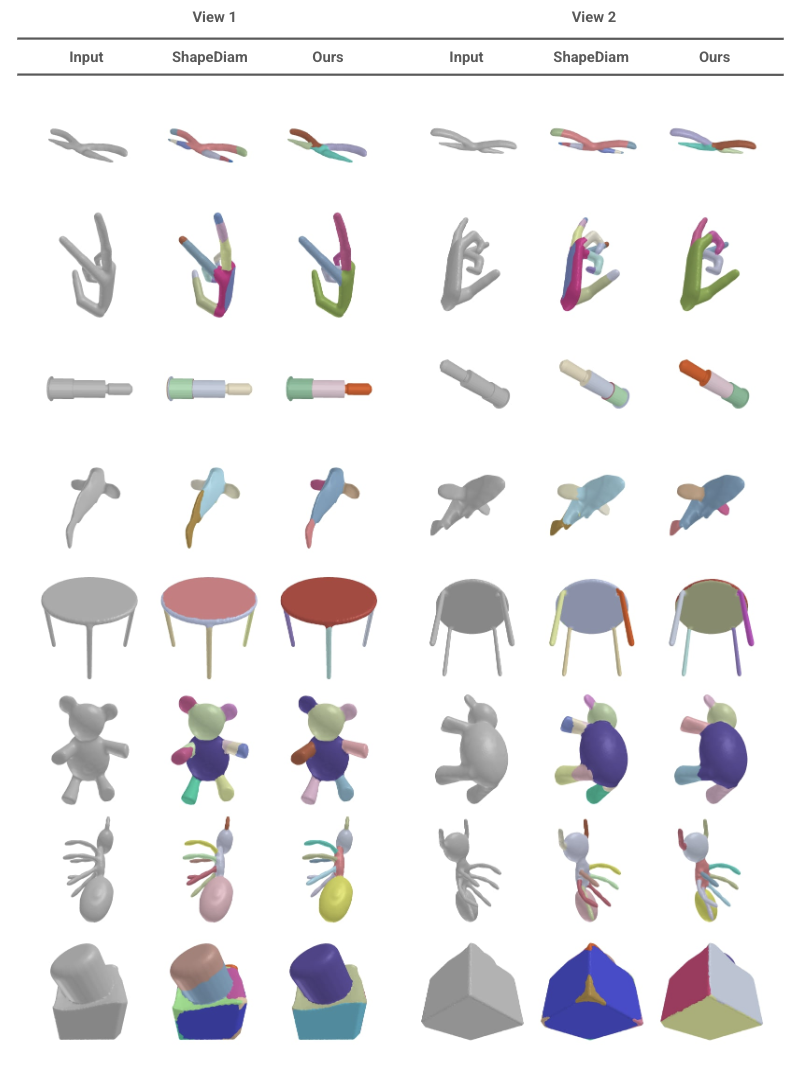}
    \caption{Additional visualizations for Segment Any Mesh vs Shape Diameter Function on the Princeton Mesh Segmentation Benchmark for automatic determination of the number of segmentation regions.}
    \label{fig:ours_vs_shapediam_princeton2}
\end{figure}

\end{document}